\documentclass[conference]{IEEEtran}

\usepackage{amssymb}
\usepackage{amsmath}
\usepackage{dsfont}
\usepackage{booktabs} 

\usepackage{graphicx,wrapfig}
\usepackage{caption}

\usepackage[nocompress]{cite}
\usepackage{url}
\usepackage{multirow}
\usepackage{float}
\usepackage{mathtools}
\usepackage[ruled, vlined,algo2e,linesnumbered]{algorithm2e}

\usepackage{tikz}
\usepackage{xcolor}
\usepackage{tkz-euclide,subfigure}
\usetikzlibrary{shapes}
\newlength{\R}
\newlength{\Z}
\newlength{\sR}
\usepackage{pgfplots}
\usepackage{soul}
\pgfplotsset{compat=newest}
\usetikzlibrary{shapes.arrows}
\usetikzlibrary{arrows.meta}

\newcommand\footnoteref[1]{\protected@xdef\@thefnmark{\ref{#1}}\@footnotemark}
\newcommand{\RN}[1]{%
	\textup{\uppercase\expandafter{\romannumeral#1}}%
}
\newcommand{\diag}{\mathop{\mathrm{diag}}}

\DeclareMathAlphabet{\mathcal}{OMS}{cmsy}{m}{n}

\newtheorem{theorem}{Theorem}

\newtheorem{proof}{Proof}

\usepackage{flushend}
\usepackage{balance}
\begin{document}

\sloppy

\title{Graph Neural Diffusion Networks for Semi-supervised Learning}


\author{\IEEEauthorblockN{Wei Ye$^\ast$, Zexi Huang$^\dagger$, Yunqi Hong$^\ast$, and Ambuj Singh$^\dagger$}
	\IEEEauthorblockA{\textit{$^\ast$ Tongji University, Shanghai 201804, China}\\
		\textit{$^\dagger$ University of California, Santa Barbara, CA 93106, USA}} 
	\{yew, wendyhong\_hyq\}@tongji.edu.cn\\
	\{zexi\_huang, ambuj\}@cs.ucsb.edu
}

\maketitle

\begin{abstract}
	Graph Convolutional Networks (GCN) is a pioneering model for graph-based semi-supervised learning. However, GCN does not perform well on sparsely-labeled graphs. Its two-layer version cannot effectively propagate the label information to the whole graph structure (i.e., the under-smoothing problem) while its deep version over-smoothens and is hard to train (i.e., the over-smoothing problem). To solve these two issues, we propose a new graph neural network called GND-Nets (for Graph Neural Diffusion Networks) that exploits the local and global neighborhood information of a vertex in a single layer. Exploiting the shallow network mitigates the over-smoothing problem while exploiting the local and global neighborhood information mitigates the under-smoothing problem. The utilization of the local and global neighborhood information of a vertex is achieved by a new graph diffusion method called neural diffusions, which integrate neural networks into the conventional linear and nonlinear graph diffusions. The adoption of neural networks makes neural diffusions adaptable to different datasets. Extensive experiments on various sparsely-labeled graphs verify the effectiveness and efficiency of GND-Nets compared to state-of-the-art approaches.
\end{abstract}

\IEEEpeerreviewmaketitle

\section{Introduction}\label{intro}

\IEEEPARstart{G}{raph} is a kind of data structure that can model the relationships between objects and thus is ubiquitous in the real world. For example, social networks, n-body systems, protein-protein interaction networks, and molecules can be modeled as graphs. Machine learning methods on graphs, such as graph-based semi-supervised learning, have received a lot of attention recently, which can be divided into two branches: shallow models and deep models.

Shallow models such as Label Spreading (LS)~\cite{DBLP:conf/nips/ZhouBLWS03} and the weighted vote Geometric Neighbor (wvGN)~\cite{ye2017learning} learning classifier exploit interdependency between vertices to infer labels. Specifically, they make use of the graph diffusion methods~\cite{andersen2006local,chung2007heat} to propagate the label information to the whole graph structure. The assumption is that the vertex labeles satisfy the principle of homophily~\cite{mcpherson2001birds}, i.e., vertices connected to each other are likely to have the same labels. Integrated with the representational power of deep neural networks, deep models such as Graph Neural Networks (GNNs)~\cite{scarselli2008graph} perform much better than shallow models. Graph Convolutional Networks (GCN)~\cite{kipf2016semi} is a pioneering GNN model that introduces a simple and well-behaved layer-wise propagation rule that is derived from the first-order approximation of spectral graph convolutions~\cite{hammond2011wavelets}. In fact, the spectral graph convolution in GCN can be considered as a low-pass filter, as pointed out by some recent works~\cite{li2018deeper,wu2019simplifying,nt2019revisiting}. For sparsely-labeled graphs, stacking too many graph convolutional layers leads to over-smoothing, i.e., vertices from different classes become indistinguishable, while stacking only few graph convolutional layers may not effectively propagate the label information to the whole graph structure, and thus leads to a low performance.

To solve the above two challenges of GCN, i.e., over-smoothing and under-smoothing, JK-Nets~\cite{xu2018representation} proposes to aggregate the output of each layer by skipping connections. It selectively exploit information from neighborhoods of different locality. Indeed, the performance of GCN is improved by aggregating the output of each layer, but not significantly (see Section~\ref{sec:classification}). One reason is that the deep GCN model with many graph convolutional layers is hard to train. Simple Graph Convolution (SGC)~\cite{wu2019simplifying} achieves similar results to GCN by removing all the nonlinear activation functions and raising the graph Laplacian matrix to the $K$-th power. However, the local information in a vertex neighborhood is not employed, which limits its performance. PPNP~\cite{klicpera2019predict} and GDC~\cite{klicpera2019diffusion} adopt the personalized PageRank diffusion~\cite{andersen2006local} or the heat kernel diffusion~\cite{chung2007heat} to aggregate the local and global neighborhood information of a vertex. The fixed weights of the graph diffusion methods are dataset-agnostic and thus may not be suitable for a specific dataset. Besides, the closed-form solution of the personalized PageRank diffusion or the heat kernel diffusion is computationally expensive.

To mitigate the two challenges of GCN,
we propose to use the local and global neighborhood information (different orders of the vertex neighborhood) in a single layer. \textbf{Exploiting the shallow network mitigates the over-smoothing problem while exploiting the local and global neighborhood information mitigates the under-smoothing problem}. Since explicitly computing each order of the vertex neighborhood information (contained in every power of the graph Laplacian matrix) is less efficient, we analyse the layer-wise propagation rule of GCN from the perspective of power iteration and derive a sequence of matrices that include the local and global neighborhood information of the graph structure. Then, we propose to aggregate the local and global neighborhood information by the Single-Layer Perceptron (SLP), which leads to a new class of graph diffusions called neural diffusions. Differing from traditional linear graph diffusions such as the personalized PageRank diffusion~\cite{andersen2006local} and the heat kernel diffusion~\cite{chung2007heat}, the weighting parameters in neural diffusions are not fixed but learned by neural networks, which makes neural diffusions adaptable to different datasets. We integrate neural diffusions into graph neural networks and develop a new GNN model called GND-Nets (for Graph Neural Diffusion Networks), which outperforms other GNNs, especially when the graph is sparsely labeled. Our main contributions can be summarized as follows:
\begin{itemize}
	\item We interpret the layer-wise propagation rule of GCN from the perspective of power iteration and propose to solve the two challenges of GCN by aggregating the local and global neighborhood information contained in a sequence of matrices generated during the convergence process of power iteration in a single layer.
	\item We propose a new class of graph diffusions called neural diffusions based on the SLP to aggregate the local and global neighborhood information. Neural diffusions are adpatable to different datasets.
	\item We integrate neural diffusions into graph neural networks and develop a new GNN model called GND-Nets. We show the effectiveness and efficiency of GND-Nets by carrying out extensive comparative studies with state-of-the-art methods on various sparsely-labeled graphs.
\end{itemize} 

\section{Preliminaries}

\subsection{Graph Convolutional Networks}
Graph Convolutional Networks (GCN)~\cite{kipf2016semi} extends convolution operations in CNNs to graphs by spectral convolutions, which is defined as follows:
\begin{equation}
\label{eqn:spectral_convolution}
g_{\boldsymbol{\theta}}(\mathbf{L})\star\mathbf{x}=g_{\boldsymbol{\theta}}(\mathbf{U}\Lambda\mathbf{U}^{\intercal})\star\mathbf{x}=\mathbf{U}g_{\boldsymbol{\theta}}(\Lambda)\mathbf{U}^{\intercal}\mathbf{x}
\end{equation}
where $\mathbf{x}\in\mathbb{R}^n$ is a signal (feature vector) on a vertex, $g_{\boldsymbol{\theta}}$ is a spectral filter~\cite{bruna2013spectral,sandryhaila2013discrete} on $\mathbf{\Lambda}$, parameterized by $\boldsymbol{\theta}\in\mathbb{R}^n$, and $\mathbf{U}^{\intercal}\mathbf{x}$ is the graph Fourier transform of signal $\mathbf{x}$.

Considering that the multiplication of matrices in Eqn.~(\ref{eqn:spectral_convolution}) has a high time complexity ($\mathcal{O}(n^2)$) and the eigendecomposion of $\mathbf{L}$ is prohibitively expensive ($\mathcal{O}(n^3)$) especially for large graphs, we can circumvent the problem by approximating $g_{\boldsymbol{\theta}}$ by a truncated expansion in terms of Chebyshev polynomials $T_k(x)$ up to the $K$-th order~\cite{hammond2011wavelets}. The Chebyshev polynomials are recursively defined as $T_k(x)=2xT_{k-1}(x)-T_{k-2}(x)$, with $T_0(x)=1$ and $T_1(x)=x$. (Please refer to~\cite{hammond2011wavelets, defferrard2016convolutional} for more details.) Thus, $g_{\boldsymbol{\theta}}$ can be approximated as follows:

\begin{equation}
\label{eqn:chebyshev}
g_{\boldsymbol{\theta}}(\mathbf{\Lambda})\simeq\sum_{k=0}^{K}\theta_kT_k(\mathbf{\widetilde{\Lambda}})
\end{equation}
where $\mathbf{\widetilde{\Lambda}}=\frac{2}{\lambda_{\mbox{max}}}\mathbf{\Lambda}-\mathbf{I}$, and $\lambda_{\mbox{max}}$ is the largest eigenvalue of $\mathbf{L}$. $\boldsymbol{\theta}\in\mathbb{R}^K$ is a vector of Chebyshev coefficients. Then, Eqn.~(\ref{eqn:spectral_convolution}) can be written as follows:
\begin{equation}
\label{eqn:chebyshev_approximate}
g_{\boldsymbol{\theta}}(\mathbf{L})\star\mathbf{x}=\sum_{k=0}^{K}\theta_kT_k(\mathbf{\widetilde{L}})\mathbf{x}
\end{equation}
where $\mathbf{\widetilde{L}}=\frac{2}{\lambda_{\mbox{max}}}\mathbf{L}-\mathbf{I}$. Eqn.~(\ref{eqn:chebyshev_approximate}) is $K$-localized, i.e., it depends only on the vertices that are maximum $K$-hop distance away from the center vertex ($K$-th order neighborhood). The time complexity of Eqn.~(\ref{eqn:spectral_convolution}) is reduced to $\mathcal{O}(e)$, where $e$ is the number of edges.

By setting $K=1$ and $\lambda_{\mbox{max}}=2$, GCN~\cite{kipf2016semi} simplifies Eqn.~(\ref{eqn:chebyshev_approximate}) as follows:
\begin{equation}
\label{eqn:simplify}
g_{\boldsymbol{\theta}}(\mathbf{L})\star\mathbf{x}\simeq\theta_0\mathbf{x} + \theta_1(\mathbf{L}-\mathbf{I})\mathbf{x}
\end{equation}
where $\theta_0$ and $\theta_1$ are two parameters.

By setting $\theta=\theta_0=-\theta_1$ and using $\mathbf{L}_{\mbox{sym}}$, Eqn.~(\ref{eqn:simplify}) can be further rewritten as:
\begin{equation}
\label{eqn:simplify_again}
g_{\boldsymbol{\theta}}(\mathbf{L})\star\mathbf{x}\simeq\left(\mathbf{I}+\mathbf{D}^{-\frac{1}{2}}\mathbf{A}\mathbf{D}^{-\frac{1}{2}}\right)\mathbf{x}\theta
\end{equation}

Since $\mathbf{I}+\mathbf{D}^{-\frac{1}{2}}\mathbf{A}\mathbf{D}^{-\frac{1}{2}}$ has eigenvalues in the range $\left[0, 2 \right] $, repeating this learning rule will cause numerical instabilities and exploding/vanishing gradients problems in deep neural networks. So GCN employs a renormalization trick $\mathbf{I}+\mathbf{D}^{-\frac{1}{2}}\mathbf{A}\mathbf{D}^{-\frac{1}{2}}\rightarrow\mathbf{\widetilde{D}}^{-\frac{1}{2}}\mathbf{\widetilde{A}}\mathbf{\widetilde{D}}^{-\frac{1}{2}}$. Then, its eigenvalues are in the range $\left[-1, 1 \right] $. Eqn. (\ref{eqn:simplify_again}) can be generalized to a signal matrix $\mathbf{X}\in\mathbb{R}^{n\times d}$ on all the vertices in a graph:
\begin{equation}
\label{eqn:generalized}
\mathbf{H}=\mathbf{\widetilde{D}}^{-\frac{1}{2}}\mathbf{\widetilde{A}}\mathbf{\widetilde{D}}^{-\frac{1}{2}}\mathbf{X}\mathbf{\Theta}
\end{equation}
where $\mathbf{\Theta}\in\mathbb{R}^{d\times r}$ is a matrix of filter parameters and $r$ is the number of filters on the vertex feature vector.

Then, the layer-wise propagation rule of GCN is defined as follows:
\begin{equation}
\label{eqn:propagation}
\mathbf{H}^{(k)}=\sigma\left(\mathbf{\widetilde{D}}^{-\frac{1}{2}}\mathbf{\widetilde{A}}\mathbf{\widetilde{D}}^{-\frac{1}{2}}\mathbf{H}^{(k-1)}\mathbf{\Theta}^{(k-1)}\right)
\end{equation}
where $\mathbf{H}^{(0)}=\mathbf{X}$, $\mathbf{\Theta}^{(k-1)}$ is the trainable filter parameter matrix in the $(k-1)$-th layer, and $\sigma(\cdot)$ is an activation function.

%

\subsection{Graph Diffusions}
Graph diffusions can be generalized as follows:
\begin{equation}
\label{eqn:grapf_diff}
\mathbf{u}^{(K)}=\sum_{k=0}^{K-1}\alpha_k\mathbf{\widetilde{W}}^k\mathbf{u}^{(0)}
\end{equation}
where $\mathbf{u}^{(0)}$ is a vector of length $n$ (the number of vertices), each entry of which denotes the initial material at each vertex. $\alpha_k$ is non-negative, which satisfies $\sum_k\alpha_k=1$ and functions as a decaying weight to ensure that the diffusion dissipates. $\mathbf{u}^{(K)}$ captures how the material diffuses over the edges of the graph. If $\alpha_k$ takes the form of $\alpha_k=(1-\gamma)\cdot\gamma^k$ with teleport probability $\gamma\in (0, 1)$, Eqn. (\ref{eqn:grapf_diff}) becomes the personalized PageRank diffusion; if $\alpha_k$ takes the form of $\alpha_k=\exp(-t)\frac{t^k}{k!}$ with the diffusion time $t$, Eqn. (\ref{eqn:grapf_diff}) becomes the heat kernel diffusion.

\section{Graph Neural Diffusion Networks}\label{sec:gndc}
\subsection{Local and Global Neighborhood Information}
To mitigate the two challenges of GCN, we propose to use the different orders of the vertex neighborhood information in a single layer. In the following, we will explain our method in detail. Firstly, we interpret the layer-wise propagation rule of GCN from the perspective of power iteration.

Inspired by SGC~\cite{wu2019simplifying}, we: (1) set all the intermediate nonlinear activation functions as linear ones $\sigma(x)=x$, (2) replace $\mathbf{\widetilde{D}}^{-\frac{1}{2}}\mathbf{\widetilde{A}}\mathbf{\widetilde{D}}^{-\frac{1}{2}}$ with $\mathbf{\widetilde{W}}=\mathbf{\widetilde{D}}^{-1}\mathbf{\widetilde{A}}$ (for the convenience of our derivation, the conclusion is also hold for $\mathbf{\widetilde{D}}^{-\frac{1}{2}}\mathbf{\widetilde{A}}\mathbf{\widetilde{D}}^{-\frac{1}{2}}$), and (3) reparameterize all the weight matrices into a single matrix $\mathbf{\Theta}=\prod_{i=0}^{k-1} \mathbf{\Theta}^{(i)}$. Then, Eqn. (\ref{eqn:propagation}) becomes the following:
\begin{equation}
\label{eqn:random_walk}
\mathbf{H}^{(k)}=\sigma\left(\mathbf{\widetilde{W}}^k\mathbf{X}\mathbf{\Theta}\right)
\end{equation}

$\mathbf{H}^{(0)}=\mathbf{Z}=\mathbf{X}\mathbf{\Theta}$ can be considered as computed by applying a linear layer (parameterized by $\mathbf{\Theta}$) on the vertex feature matrix $\mathbf{X}$. For each column vector $\mathbf{z}\in\mathbf{Z}$, we have the following theorem:

\begin{theorem}
	\label{tho:power}
	If the graph underlying $\mathbf{\widetilde{W}}$ is non-bipartite, the vector $\mathbf{\widetilde{W}}^k\mathbf{z}$ converges and the limit is the dominant eigenvector of $\mathbf{\widetilde{W}}$.
\end{theorem}

\begin{figure*}[!htb]
	\hspace*{\fill}
	\centering
	\subfigure[$k=0$]{\includegraphics [width=0.2\textwidth]{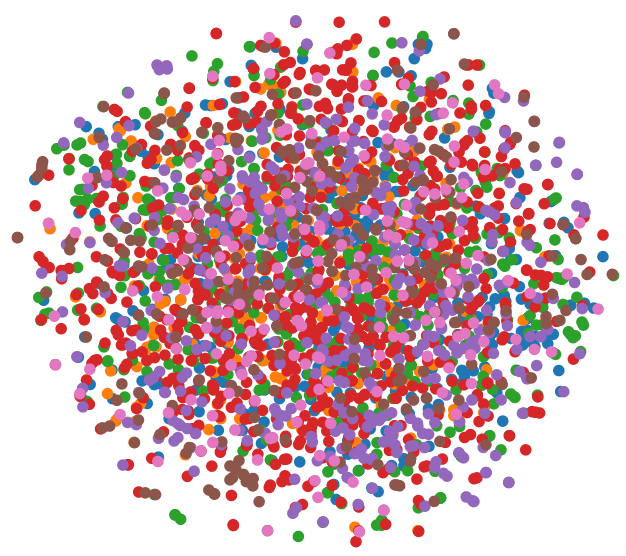}}
	\hfill
	\centering
	\subfigure[$k=1$]{\includegraphics [width=0.2\textwidth]{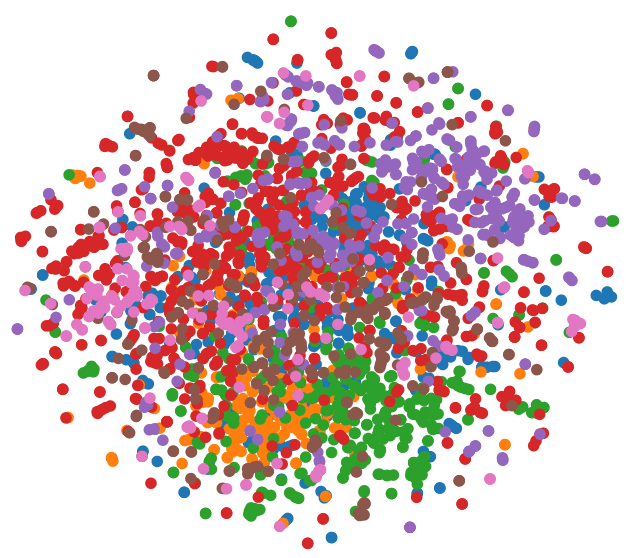}}
	\hfill
	\centering
	\subfigure[$k=19$]{\includegraphics [width=0.2\textwidth]{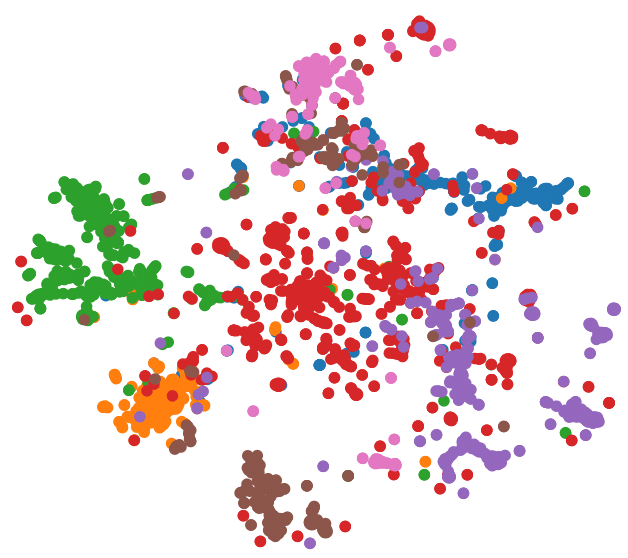}}
	\hfill
	\centering
	\subfigure[$k=10,000$]{\includegraphics [width=0.2\textwidth]{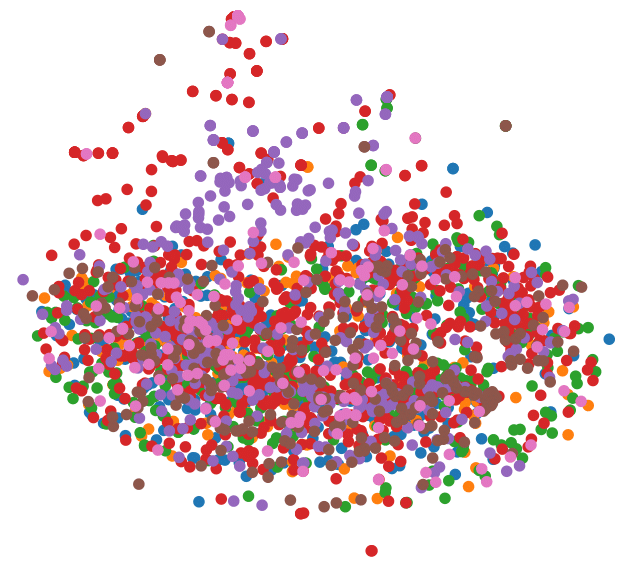}}
	\hspace*{\fill}
	
	\caption{The t-SNE visualization of the random projection of the feature matrix of dataset \textsc{Cora}. Power iteration reveals the cluster structures after $k=19$ iterations. Colors denote classes.} 
	\label{fig:random_projection}
\end{figure*}

\begin{proof}
	Power iteration (PI) is an efficient and popular method to compute the dominant eigenvector of a matrix. PI starts with an initial vector $\mathbf{v}^{0}\neq\mathbf{0}$ and iteratively updates as follows:
	\begin{equation}
	\label{fig:update}
	\mathbf{v}^k=\frac{\mathbf{\widetilde{W}}\mathbf{v}^{k-1}}{|\mathbf{\widetilde{W}}\mathbf{v}^{k-1} |_1}
	\end{equation}
	
	Suppose $\mathbf{\widetilde{W}}$ has right eigenvectors $\mathbf{U}=[\mathbf{u}_1;\mathbf{u}_2;\ldots;\mathbf{u}_n]$ with right eigenvalues $\mathbf{\Lambda}=\diag([\lambda_1,\lambda_2,\ldots,\lambda_n])$, where $\lambda_1=1$ and $\mathbf{u}_1$ is the dominant eigenvector of all ones. We have $\mathbf{\widetilde{W}}\mathbf{U}=\mathbf{\Lambda}\mathbf{U}$ and in general $\mathbf{\widetilde{W}}^k\mathbf{U}=\mathbf{\Lambda}^k\mathbf{U}$. When ignoring renormalization, Eqn. (\ref{fig:update}) can be written as:
	\begin{equation}
	\label{fig:updateT}
	\begin{split}
	\mathbf{v}^k&=\mathbf{\widetilde{W}}\mathbf{v}^{k-1}=\mathbf{\widetilde{W}}^2\mathbf{v}^{k-2}=\cdots =\mathbf{\widetilde{W}}^k\mathbf{v}^0 \\
	&=\mathbf{\widetilde{W}}^k\left( c_1\mathbf{u}_1+c_2\mathbf{u}_2+\cdots +c_n\mathbf{u}_n\right) \\
	&=c_1\lambda_1^k\mathbf{u}_1+c_2\lambda_2^k\mathbf{u}_2+\cdots +c_n\lambda_n^k\mathbf{u}_n
	\end{split}
	\end{equation}
	where $\mathbf{v}^0=\mathbf{z}$ can be denoted by $c_1\mathbf{u}_1+c_2\mathbf{u}_2+\cdots +c_n\mathbf{u}_n$, which is a linear combination of all the original orthonormal eigenvectors. Since the orthonormal eigenvectors form a basis for $\mathbb{R}^n$, any vector can be expanded by them. 
	
	According to Eqn. (\ref{fig:updateT}), we have
	\begin{equation}
	\frac{\mathbf{v}^k}{c_1\lambda_1^k}=\mathbf{u}_1+\frac{c_2}{c_1}\left(\frac{\lambda_2}{\lambda_1} \right)^k \mathbf{u}_2+\cdots +\frac{c_n}{c_1}\left(\frac{\lambda_n}{\lambda_1} \right)^k \mathbf{u}_n
	\end{equation}
	
	So the convergence rate of PI towards the dominant eigenvector $\mathbf{u}_1$ depends on the significant terms $\left(\frac{\lambda_i}{\lambda_1} \right)^k (2\leqslant i \leqslant n)$. Note that $\lambda_1$ is the largest eigenvalue that makes the significant terms less than 1. If letting PI run long enough ($k$ is a large number), it will converge to the dominant eigenvector $\mathbf{u}_1$.
\end{proof}

Theorem~\ref{tho:power} indicates that if $k$ is very large and $\lambda_1>\lambda_2>\cdots>\lambda_n$, each feature column of $\mathbf{\widetilde{W}}^k\mathbf{X}\mathbf{\Theta}$ will converge to the dominant eigenvector $\mathbf{u}_1$ of $\mathbf{\widetilde{W}}$ regardless of $\mathbf{X}$ and $\mathbf{\Theta}$, which is of little use in classification. This is the reason why the performance of the deep-layer GCN deteriorates on some datasets. However, the intermediate vectors generated by PI during the convergence process can be very useful. 

For example, as shown in Figure~\ref{fig:random_projection}(a)--(d), we first use the method described in~\cite{glorot2010understanding} to randomly initialize the weights of $\mathbf{\Theta}$ (the number of filters $r$ is set to 16), and then project $\mathbf{\widetilde{W}}^k\mathbf{X}\mathbf{\Theta}$ of dataset \textsc{Cora} to the 2D space by t-SNE~\cite{van2008visualizing}. Figure~\ref{fig:random_projection}(a) shows that initially we have no cluster structures in the data. Even after one iteration (as shown in Figure~\ref{fig:random_projection}(b), data points start to form clusters. After 19 iterations as shown in Figure~\ref{fig:random_projection}(c), we can see obvious cluster structures in the data. Figure~\ref{fig:random_projection}(d) indicates that t-SNE cannot separate the different clusters after 10,000 power iteraions, because all the data points converge to the dominant eigenvector of all ones. Note that in this process, no label information is used to guide the learning.
If the Laplacian matrix of the graph structure captures the pairwise vertex similarities, i.e., the graph satisfies the principle of homophily, power iteration will make cluster separated and the provided label information will just accelerate this process.


\subsection{Neural Diffusions}
GCN uses only one iteration of power iteration ($k=1$) that is not sufficient to propagate the label information to the whole graph structure when the number of the labeled vertices is scarce. 
We use power iteration $k=K$ times to generate a sequence of intermediate matrices $\{\mathbf{Z},\mathbf{\widetilde{W}}\mathbf{Z},\ldots,\mathbf{\widetilde{W}}^{K-1}\mathbf{Z}\}$ ($\mathbf{Z}=\mathbf{X}\mathbf{\Theta}$). We propose to aggregate all the local and global neighborhood information contained in these matrices in a single layer for semi-supervised classification on sparsely-labeled graphs. The aggregation is achieved by neural networks such as the Single-Layer Perceptron (SLP). 

The aggregation by SLP is as follows:
\begin{equation}
\begin{split}
\mathbf{H}^{(K)}&=\sigma\left(\alpha_0\mathbf{Z}+\alpha_1\mathbf{\widetilde{W}}\mathbf{Z}+\cdots+\alpha_{K-1}\mathbf{\widetilde{W}}^{K-1}\mathbf{Z}\right)\\
&=\sigma\left(\left(\alpha_0\mathbf{I}+\alpha_1\mathbf{\widetilde{W}}+\cdots+\alpha_{K-1}\mathbf{\widetilde{W}}^{K-1}\right)\mathbf{Z}\right)\\
&=\sigma\left(\left(\sum_{k=0}^{K-1}\alpha_k\mathbf{\widetilde{W}}^k\right)\mathbf{Z}\right)
\end{split}
\label{eqn:agg}
\end{equation}
where $\alpha_k$ ($0\leq k\leq K-1$) is the weighting parameters of the SLP.

Compared with Eqn. (\ref{eqn:grapf_diff}), the term in the inner parentheses is a truncated graph diffusion. By relaxing the constraint $\sum_k\alpha_k=1$, allowing $\alpha_k$ to be arbitrary values and letting the SLP adaptively learn them, we arrive at a new graph diffusion method: \textbf{neural diffusions}. Note that for implementations, we first flatten $\mathbf{\widetilde{W}}^k\mathbf{Z}$ ($0\leq k\leq K-1$) into vectors and consider dimension hops as attribute. Then, we use the SLP to aggregate all these $K$ vectors. Because the number of the filters of the SLP are set to one, we need to reshape the outputs of the SLP by $f^{-1}\left(\cdot\right)$ into a matrix $\mathbf{H}^{(K)}\in\mathbb{R}^{n\times r}$, which has the same dimension as $\mathbf{Z}$. Eqn. (\ref{eqn:agg}) is derived from Eqn. (\ref{eqn:grapf_diff}), which is a linear graph diffusion. 

\section{Experimental Evaluation}\label{sec:experiments}
\subsection{Experimental Setup}\label{sec:setup}
We compare GND-Nets with several state-of-the-art GNN models, including GCN~\cite{kipf2016semi}, Geom-GCN~\cite{pei2020geom}, ChebyNet~\cite{defferrard2016convolutional}, JK-Nets~\cite{xu2018representation} with max-pooling strategy for neighborhood aggregation, SGC~\cite{wu2019simplifying}, PPNP~\cite{klicpera2019predict}, PPNP-HK (the personalized PageRank diffusion is replaced by the heat kernel diffusion), N-GCN~\cite{abu2018n}, MixHop~\cite{abu2019mixhop}, LanczosNet~\cite{liao2019lanczosnet}, and DCNN~\cite{atwood2016diffusion}. The hyperparameters of all the baselines are set as in their original papers. For JK-Nets, we also tune the number of layers from one to 20 and report the best results. For a fair comparison, we run each method at most 1,000 epochs. We conduct experiments on the benchmark datasets to evaluate the classification performance of GND-Nets and its baselines. The statistics of the benchmark datasets used in this paper are shown in Table~\ref{tab:statistics}. We make our code publicly available at Github\footnote{\url{https://github.com/yeweiysh/GND-Nets}}.

\begin{table*}[!htb]
	\small
	\centering
	\caption{Statistics of datasets.}
	\label{tab:statistics}
	\begin{tabular}{lllllll}
		\toprule
		Dataset                 &\textsc{Cora}       &\textsc{Cora-ML} &\textsc{Citeseer}  &\textsc{Pubmed}   &\textsc{Amazon Computers}&\textsc{Amazon Photo}  \\
		\midrule
		\# Vertices             &2,708       &2,810       &3,327        &19,717        &13,381                      &7,487                   \\ 
		\# Edges                  &5,429      &7,981       &4,732        &44,338      &245,778                   &119,043               \\ 
		\# Features              &1,433      &2,879       &3,703       &500            &767                           &745                       \\ 
		\# Classes                &7              &7               &6                &3                 &10                             &8                           \\
		\bottomrule
	\end{tabular}
\end{table*}


For GND-Nets, we set the hyperparameters as follows: We use Adam~\cite{kingma2015adam} optimization method with a learning rate of 0.005, L2 regularization factor of $5\times10^{-4}$ for each layer. The window size is set to 50, i.e., we terminate training if the validation loss does not decrease for 50 consecutive epochs. Weights in each layer are initialized according to the initialization method described in~\cite{glorot2010understanding}. The number of filters of $\mathbf{\Theta}$ for GND-Nets is fixed to 16 for each dataset. The number of filters in each layer equals to the number of classes in the datasets. Dropout rate is set to 0.6. We run all the experiments on a machine with a dual-core Intel(R) Xeon(R) E5-2678 CPU@2.50GHz, 128 GB memory, and an Nvidia GeForce RTX 2080 Ti GPU. 




\subsection{Results}
\subsubsection{Classification}\label{sec:classification}
Tables~\ref{tab:cora}--\ref{tab:amazonc} show the average classification accuracy of each method on benchmark datasets over 30 different data splits. We can see that the performance of each method increases dramatically with the increasing number of the labeled vertices in graphs such as \textsc{Cora}, \textsc{Citeseer}, and \textsc{Pubmed}. On \textsc{Cora-ML}, 
the performances of ChebyNet, MixHop, and LanczosNet do not increase much. On \textsc{Amazon Computers} and \textsc{Amazon Photo}, all the comparison methods do not perform well and are defeated by GND-Nets by a large margin. Note that GND-Nets achieves the best results in most of the cases. Specifically, GND-Nets performs the best in 4 out of 30 cases. GND-Nets performs significantly better than GCN on all the datasets, which suggests that aggregating the local and global neighborhood information by neural diffusions in a single layer does improve the performance when the graph is sparsely-labeled. GCN uses only the second-order neighborhood information and performs worse than many other methods, because the vertex label information does not propagate well into the graph structure. 

By aggregating the output of each layer of GCN, JK-Nets improves the performance of GCN in most cases. On \textsc{Pubmed}, \textsc{Amazon Computers}, and \textsc{Amazon Photo}, PPNP and PPNP-HK run out of memory. 
We can see that GND-Nets is superior to PPNP, PPNP-HK, and DCNN on all the datasets. This illustrates the strength of adaptive aggregation weights for different orders of neighborhood over fixed aggregation weights. In addition, GND-Nets is also better than the three multi-scale graph convolution methods N-GCN, MixHop, and LanczosNet. SGC only uses the higher-order neighborhood information and is defeated by GND-Nets. Geom-GCN exploits the three node embedding methods to capture the structural information and long-range dependencies between vertices in graphs. The strategy is very complicated and ineffective on large datasets. 

\begin{table*}[!htb]
	\centering
	\caption{Average classification accuracy (\%) over 30 different data splits on \textsc{Cora} with varying sizes of labeled vertices. \# of LV/class is the abbreviation for the number of labeled vertices/class.}
	\label{tab:cora}
	\begin{tabular}{llllll}
		\toprule
		\# of LV/class          &1                              &2                          &3                       &4                   &5       \\\hline
		GND-Nets           &56.04$\pm$11.54      &66.15$\pm$8.58                  &70.52$\pm$6.38               &73.81$\pm$3.62            &74.79$\pm$3.15   \\
		GCN                       &36.33$\pm$14.90        &52.03$\pm$16.63       &60.71$\pm$10.90        &66.24$\pm$4.57          &68.61$\pm$3.89\\ 
		Geom-GCN          &20.59$\pm$3.57         &23.46$\pm$2.97         &24.93$\pm$2.86        &27.64$\pm$2.43          &29.87$\pm$2.91   \\
		ChebyNet             &24.21$\pm$9.47           &32.63$\pm$13.31        &39.18$\pm$13.49       &47.78$\pm$11.80         &55.38$\pm$8.67   \\
		JK-Nets                &43.67$\pm$8.66         &55.65$\pm$7.15           &60.61$\pm$6.19          &64.58$\pm$4.47           &66.99$\pm$4.29    \\
		SGC                       &38.97$\pm$11.81         &55.59$\pm$8.31           &61.31$\pm$6.73          &65.50$\pm$4.49           &67.44$\pm$3.75    \\
		PPNP                     &45.91$\pm$12.46         &59.47$\pm$9.21           &65.66$\pm$7.73         &69.48$\pm$5.86          &71.43$\pm$5.08   \\
		PPNP-HK                &35.99$\pm$12.17                     &57.64$\pm$12.26                     &64.78$\pm$7.81             &69.07$\pm$4.91            &71.27$\pm$4.56 \\
		N-GCN                  &42.03$\pm$11.14                     &53.06$\pm$7.48                    &58.09$\pm$5.14             &61.98$\pm$4.16             &64.21$\pm$3.37\\
		MixHop                  &31.09$\pm$13.46                    &44.50$\pm$11.61                   &52.41$\pm$8.99                &58.38$\pm$8.51            &62.07$\pm$8.03\\
		LanczosNet          &44.34$\pm$10.46                   &55.71$\pm$6.88                     &61.01$\pm$5.96                &64.75$\pm$4.10             &66.72$\pm$4.63\\
		DCNN                      &21.52$\pm$9.04          &26.55$\pm$10.54   &36.90$\pm$9.34                 &44.72$\pm$5.97             &49.19$\pm$4.13\\
		\bottomrule
	\end{tabular}
\end{table*}

\begin{table*}[!htb]
	\centering
	\caption{Average classification accuracy (\%) over 30 different data splits on \textsc{Cora-ML} with varying sizes of labeled vertices. N/A means the results are not available.}
	\label{tab:citeseer}
	\begin{tabular}{llllll}
		\toprule
		\# of LV/class          &1                              &2                          &3                       &4                   &5       \\\hline
		GND-Nets           &43.84$\pm$8.36                     &52.76$\pm$6.73                  &58.76$\pm$5.37               &61.84$\pm$4.16            &63.66$\pm$3.98   \\
		GCN                          &18.59$\pm$7.35                      &23.42$\pm$9.24                 &23.89$\pm$9.99               &24.55$\pm$9.71           &28.09$\pm$11.81\\ 
		Geom-GCN              &N/A                     &N/A                  &N/A              &N/A           &N/A   \\
		ChebyNet                 &22.67$\pm$4.43                      &23.88$\pm$4.27                  &24.22$\pm$3.94               &25.65$\pm$4.11           &26.77$\pm$5.16   \\
		JK-Nets                    &22.07$\pm$8.83                     &26.51$\pm$10.20                   &27.03$\pm$11.67              &31.79$\pm$15.65           &34.22$\pm$15.54    \\
		SGC                           &27.25$\pm$8.40                    &34.79$\pm$8.87                    &41.47$\pm$9.65               &46.72$\pm$9.03           &52.06$\pm$7.39    \\
		PPNP                        &20.47$\pm$8.59                     &27.47$\pm$11.21                    &28.85$\pm$11.86             &32.57$\pm$12.18          &39.80$\pm$11.75   \\
		PPNP-HK                &17.86$\pm$7.05                     &22.27$\pm$7.99                     &22.58$\pm$9.08             &23.95$\pm$10.15            &25.17$\pm$10.12 \\
		N-GCN                     &22.92$\pm$7.69                     &29.22$\pm$7.60                    &31.71$\pm$7.43             &35.28$\pm$7.42             &38.33$\pm$9.11\\
		MixHop                     &20.07$\pm$6.19                     &22.34$\pm$6.73                    &22.77$\pm$6.89             &23.99$\pm$7.08             &26.08$\pm$7.67\\
		LanczosNet              &13.66$\pm$2.90                    &13.00$\pm$2.48                   &13.17$\pm$2.20                &13.33$\pm$2.41             &13.32$\pm$4.06\\
		DCNN                         &21.73$\pm$8.89                    &26.08$\pm$10.74                 &28.08$\pm$12.52             &33.59$\pm$15.14        &43.36$\pm$13.70\\
		\bottomrule
	\end{tabular}
\end{table*}

\begin{table*}[!htb]
	\centering
	\caption{Average classification accuracy (\%) over 30 different data splits on \textsc{Citeseer} with varying sizes of labeled vertices.}
	\label{tab:cora_ml}
	\begin{tabular}{llllll}
		\toprule
		\# of LV/class          &1                              &2                          &3                       &4                   &5       \\\hline
		GND-Nets           &39.33$\pm$10.60      &50.27$\pm$6.98                  &55.52$\pm$4.84               &58.46$\pm$4.37            &59.63$\pm$4.97   \\
		GCN                          &25.80$\pm$7.37                      &32.45$\pm$9.95                 &38.66$\pm$11.56               &45.78$\pm$13.56           &46.43$\pm$13.09\\ 
		Geom-GCN              &21.71$\pm$3.50                       &24.33$\pm$4.82                  &27.33$\pm$5.89              &30.49$\pm$7.24           &35.50$\pm$8.40   \\
		ChebyNet                 &23.09$\pm$5.68                      &27.82$\pm$9.93                  &30.84$\pm$12.78              &34.27$\pm$15.71           &38.79$\pm$17.29   \\
		JK-Nets                    &35.65$\pm$8.10                     &46.48$\pm$5.14                   &52.51$\pm$4.89              &54.59$\pm$4.99            &56.03$\pm$4.41    \\
		SGC                           &28.43$\pm$6.91                     &33.70$\pm$9.57                    &38.67$\pm$12.11             &46.37$\pm$11.71            &51.67$\pm$13.31    \\
		PPNP                        &35.55$\pm$11.62                     &47.02$\pm$8.49                    &53.28$\pm$5.31              &57.12$\pm$5.35             &59.32$\pm$3.78   \\
		PPNP-HK                &31.36$\pm$6.35                       &41.04$\pm$8.08                     &48.42$\pm$6.08             &53.10$\pm$4.94             &56.24$\pm$4.56 \\
		N-GCN                     &28.12$\pm$8.18                       &30.75$\pm$9.59                   &35.31$\pm$10.00               &41.03$\pm$10.96             &42.26$\pm$9.80\\
		MixHop                     &31.82$\pm$8.27                      &41.66$\pm$7.78                    &48.21$\pm$6.05               &51.33$\pm$8.90             &54.91$\pm$5.02\\
		LanczosNet             &31.90$\pm$9.32                       &40.41$\pm$6.76                   &45.88$\pm$6.25              &48.99$\pm$5.47             &51.53$\pm$4.36\\
		DCNN                        &27.32$\pm$7.35                       &34.14$\pm$7.45                   &39.85$\pm$6.84              &44.37$\pm$7.57              &48.26$\pm$5.46\\
		\bottomrule
	\end{tabular}
\end{table*}

\begin{table*}[!htb]
	\centering
	\caption{Average classification accuracy (\%) over 30 different data splits on \textsc{Pubmed} with varying sizes of labeled vertices.}
	\label{tab:pubmed}
	\begin{tabular}{llllll}
		\toprule
		\# of LV/class            &1                       &2                  &3                            &4                &5       \\\hline
		GND-Nets             &58.63$\pm$9.83               &65.52$\pm$9.57         &68.12$\pm$5.94     &70.43$\pm$4.87        &71.10$\pm$4.48   \\
		GCN                         &47.34$\pm$11.72        &58.66$\pm$9.62           &62.40$\pm$7.66        &66.02$\pm$5.43       &67.47$\pm$4.40 \\ 
		Geom-GCN             &N/A                                 &N/A                                    &N/A                                &N/A                               &N/A    \\
		ChebyNet                 &45.95$\pm$8.56         &52.81$\pm$10.94         &58.26$\pm$8.02        &60.57$\pm$8.19         &62.31$\pm$6.92   \\
		JK-Nets                   &49.38$\pm$11.05        &59.02$\pm$9.45           &62.94$\pm$7.41         &65.25$\pm$5.56         &66.46$\pm$5.78    \\
		SGC                          &53.18$\pm$10.00         &60.68$\pm$7.38           &64.29$\pm$5.20          &66.69$\pm$4.59           &67.56$\pm$4.18    \\
		PPNP                        &N/A                                 &N/A                                    &N/A                                &N/A                                 &N/A    \\
		PPNP-HK                        &N/A                                 &N/A                                    &N/A                                &N/A                                 &N/A    \\
		N-GCN                     &51.37$\pm$10.02                     &58.89$\pm$8.27          &62.81$\pm$4.42                    &64.81$\pm$4.73           &66.17$\pm$4.30\\
		MixHop                     &44.50$\pm$13.84                    &48.11$\pm$13.66            &55.73$\pm$9.31                  &60.02$\pm$7.03           &62.16$\pm$6.83\\
		LanczosNet              &52.21$\pm$10.12                    &60.55$\pm$10.51           &65.00$\pm$5.59                  &67.51$\pm$4.72            &68.27$\pm$4.56\\
		DCNN                         &49.61$\pm$7.83                    &58.01$\pm$7.82              &61.01$\pm$6.63                   &63.43$\pm$5.26           &65.49$\pm$4.73\\
		\bottomrule
	\end{tabular}
\end{table*}

\begin{table*}[!htb]
	\centering
	\caption{Average classification accuracy (\%) over 30 different data splits on \textsc{Amazon Computers} with varying sizes of labeled vertices.}
	\label{tab:amazonp}
	\begin{tabular}{llllll}
		\toprule
		\# of LV/class          &1                              &2                          &3                       &4                     &5       \\\hline
		GND-Nets           &33.48$\pm$12.87                     &42.82$\pm$15.27                 &47.71$\pm$17.82               &50.40$\pm$20.28                 &58.52$\pm$16.55   \\
		GCN                          &10.54$\pm$8.95                      &12.64$\pm$11.12                 &13.41$\pm$12.75              &12.13$\pm$10.89           &13.30$\pm$14.29\\ 
		Geom-GCN              &N/A                     &N/A                  &N/A              &N/A           &N/A   \\
		ChebyNet                 &12.50$\pm$5.66                      &11.84$\pm$5.39                  &11.68$\pm$4.92               &11.49$\pm$4.96           &13.06$\pm$5.04   \\
		JK-Nets                    &8.31$\pm$8.81                     &9.29$\pm$7.96                   &12.36$\pm$9.42              &10.71$\pm$9.14           &10.91$\pm$8.10    \\
		SGC                           &10.66$\pm$10.66                    &13.49$\pm$10.76                    &12.38$\pm$11.23               &13.43$\pm$11.82          &13.25$\pm$12.84    \\
        PPNP                        &N/A                                 &N/A                                    &N/A                                &N/A                                 &N/A    \\
        PPNP-HK                        &N/A                                 &N/A                                    &N/A                                &N/A                                 &N/A    \\
		N-GCN                     &13.84$\pm$10.85                     &15.57$\pm$7.84                    &16.52$\pm$7.21             &18.75$\pm$8.07             &19.35$\pm$7.45\\
		MixHop                     &10.27$\pm$7.82                     &10.91$\pm$8.16                    &10.34$\pm$7.17             &11.57$\pm$9.12             &10.95$\pm$6.71\\
		LanczosNet              &9.49$\pm$6.59                    &9.15$\pm$6.99                   &9.13$\pm$6.74               &7.69$\pm$4.63             &7.01$\pm$4.14\\
		DCNN                         &9.35$\pm$9.73                    &12.25$\pm$11.88                &13.84$\pm$14.36            &13.47$\pm$14.92        &17.28$\pm$18.71\\
		\bottomrule
	\end{tabular}
\end{table*}

\begin{table*}[!htb]
	\centering
	\caption{Average classification accuracy (\%) over 30 different data splits on \textsc{Amazon Photo} with varying sizes of labeled vertices.}
	\label{tab:amazonc}
	\begin{tabular}{llllll}
		\toprule
		\# of LV/class          &1                              &2                          &3                       &4                   &5       \\\hline
		GND-Nets           &55.98$\pm$13.50        &68.24$\pm$5.23                 &70.32$\pm$6.08               &72.85$\pm$6.17            &75.50$\pm$5.07   \\
		GCN                          &14.52$\pm$8.60                      &13.68$\pm$6.47                 &16.11$\pm$6.84              &15.01$\pm$6.96           &17.48$\pm$10.87\\ 
		Geom-GCN              &N/A                     &N/A                  &N/A              &N/A           &N/A   \\
		ChebyNet                 &17.06$\pm$4.17                      &17.37$\pm$4.42                  &18.34$\pm$5.07               &18.89$\pm$4.45           &19.23$\pm$4.91   \\
		JK-Nets                    &15.33$\pm$9.34                     &17.90$\pm$9.79                   &17.13$\pm$9.85              &21.34$\pm$13.44           &23.70$\pm$13.18    \\
		SGC                           &14.63$\pm$7.32                    &12.72$\pm$5.85                    &16.11$\pm$7.35               &15.94$\pm$7.70          &17.00$\pm$10.99    \\
        PPNP                        &N/A                                 &N/A                                    &N/A                                &N/A                                 &N/A    \\
        PPNP-HK                        &N/A                                 &N/A                                    &N/A                                &N/A                                 &N/A    \\
		N-GCN                     &15.04$\pm$5.21                     &15.96$\pm$5.18                    &18.64$\pm$6.27             &20.81$\pm$5.17             &24.15$\pm$7.75\\
		MixHop                     &12.30$\pm$4.07                     &12.79$\pm$4.13                    &13.68$\pm$4.11             &15.03$\pm$4.42             &16.30$\pm$7.03\\
		LanczosNet              &8.85$\pm$4.91                    &10.55$\pm$4.13                   &11.40$\pm$4.71               &10.95$\pm$5.48             &12.21$\pm$5.86\\
		DCNN                         &15.19$\pm$5.79                  &16.80$\pm$8.59                   &21.07$\pm$11.65            &29.60$\pm$20.26           &37.66$\pm$23.90\\
		\bottomrule
	\end{tabular}
\end{table*}

\section{Conclusion}\label{conclusion}
In this work, we have developed a new GNN model called GND-Nets. It performs well especially when the number of the labeled vertices in a graph is scarce. GND-Nets is proposed to mitigate the two drawbacks (the under-smoothing and over-smoothing problems) of GCN, i.e., its two-layer version cannot effectively propagate the label information to the whole graph structure while its deep version is hard to train and usually underperforms. GND-Nets exploits the strategy of using the local and global neighborhood information of a vertex in a single layer to mitigate the two drawbacks of GCN. The utilization of the local and global neighborhood information is performed by the proposed neural diffusions that integrate neural networks into the traditional linear and nonlinear graph diffusions. GND-Nets effectively and efficiently outperforms state-of-the-art competing methods. In the future, we would like to extend neural diffusions to include other advanced architectures such as RNN and Transformer.

\bibliographystyle{IEEEtran}
\bibliography{reference}

\end{document}